\documentclass[10pt,twocolumn,letterpaper]{article}

\usepackage{cvpr}              

\usepackage{marvosym}
\usepackage{algorithm}
\usepackage{algorithmic}
\usepackage{multirow}
\usepackage{tablefootnote}
\usepackage{float}
\usepackage{stfloats}

\usepackage{amsbsy} 
\usepackage{wasysym} 

\definecolor{cvprblue}{rgb}{0.21,0.49,0.74}
\usepackage[pagebackref,breaklinks,colorlinks,allcolors=cvprblue]{hyperref}

\def\paperID{2} 
\def\confName{CVPR}
\def\confYear{2025}

\title{ZFusion: An Effective Fuser of Camera and 4D Radar for 3D Object Perception in Autonomous Driving}

\author{
  Sheng Yang \textsuperscript{1\textdagger} \quad 
  Tong Zhan\textsuperscript{1\textdagger} \quad 
  Shichen Qiao\textsuperscript{1\textdagger} \quad 
  Jicheng Gong\textsuperscript{2} \quad 
  Qing Yang\textsuperscript{2} \\
  Jian Wang\textsuperscript{1 \Letter{}} \quad
  Yanfeng Lu\textsuperscript{3 \Letter{}} \\
  \small{\textdagger Equal contribution \quad \Letter{} Corresponding authors} \\
  \textsuperscript{1} School of Data Science, Fudan University \\
  \textsuperscript{2} ZF (China) Investment Co., Ltd. \\
  \textsuperscript{3} Institute of Automation, Chinese Academy of Sciences
}

\begin{document}
\maketitle

\begin{abstract}
Reliable 3D object perception is essential in autonomous driving. Owing to its sensing capabilities in all weather conditions, 4D radar has recently received much attention. However, compared to LiDAR, 4D radar provides much sparser point cloud. In this paper, we propose a 3D object detection method, termed ZFusion, which fuses 4D radar and vision modality. As the core of ZFusion, our proposed FP-DDCA (Feature Pyramid-Double Deformable Cross Attention) fuser complements the (sparse) radar information and (dense) vision information, effectively. Specifically, with a feature-pyramid structure, the FP-DDCA fuser packs Transformer blocks to interactively fuse multi-modal features at different scales, thus enhancing perception accuracy. In addition, we utilize the Depth-Context-Split view transformation module due to the physical properties of 4D radar. Considering that 4D radar has a much lower cost than LiDAR, ZFusion is an attractive alternative to LiDAR-based methods. In typical traffic scenarios like the VoD (View-of-Delft) dataset, experiments show that with reasonable inference speed, ZFusion achieved the state-of-the-art mAP (mean average precision) in the region of interest, while having competitive mAP in the entire area compared to the baseline methods, which demonstrates performance close to LiDAR and greatly outperforms those camera-only methods.
\end{abstract}
\section{Introduction}
\label{sec:intro}

Reliable object perception modules are the cornerstone of autonomous driving~\cite{2024tiv, Han20234DMR, radarcamerasurvey}. The main task of the perception module is to accurately capture the geometric information of objects, including relative position to the car, size, etc. 2D object detection methods such as YOLO~\cite{yolo2015, reis2024realtime} often fail to collect the relative position (depth) of the interested objects.
To overcome this issue, 3D object perception methods~\cite{li2022bevformer, philion2020lift} estimate the relative position of objects by mapping vision features to the bird's-eye view (BEV) space, thereby enabling more accurate perception.

\begin{figure}[t]
    \begin{center}
    \centerline{\includegraphics[width=\linewidth]{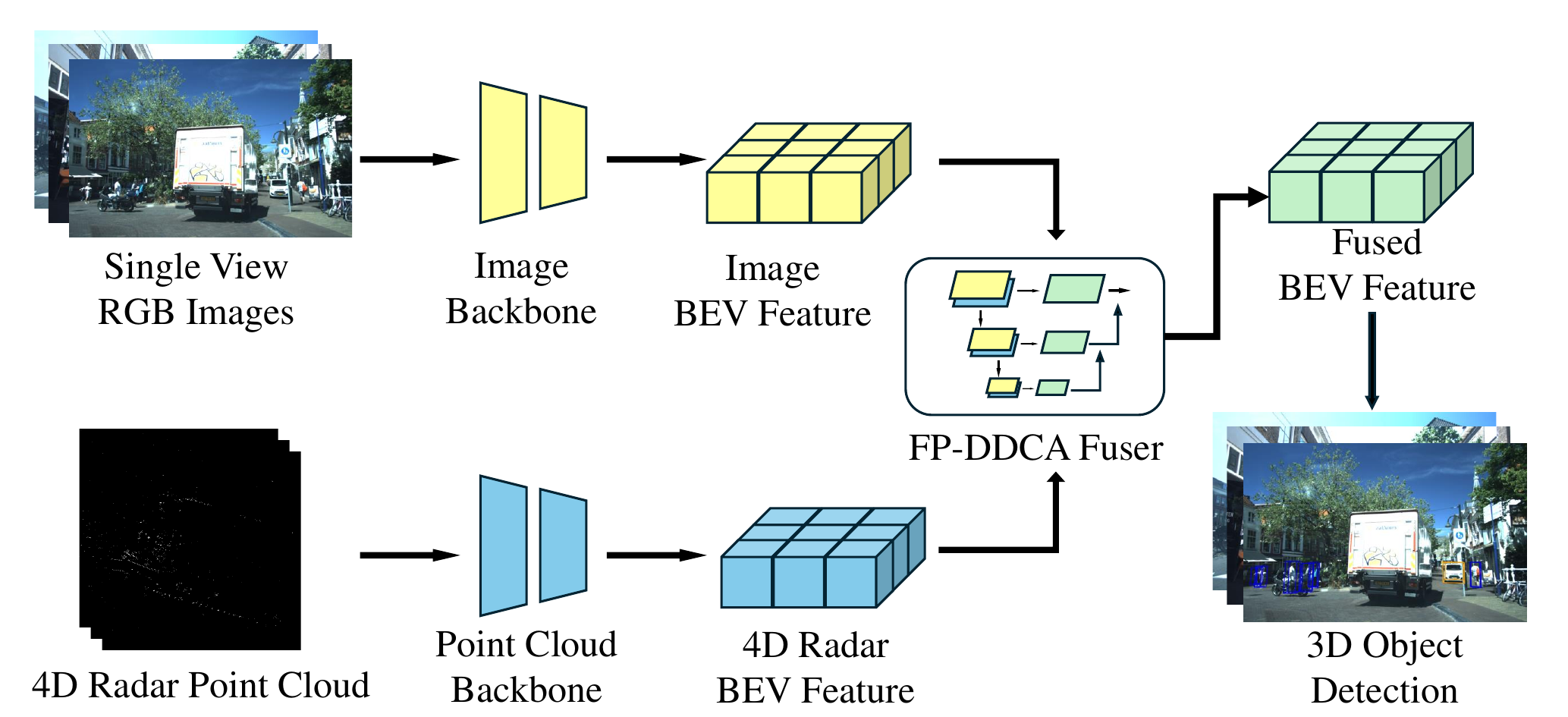}}
    \caption{ZFusion pipeline based on FP-DDCA fuser.}
    \label{fig:pipeline}
    \end{center}
\end{figure}




Reconstructing a 3D scene from 2D input is ill-posed and is faced with pure-vision schemes. Different types of sensors (e.g., LiDAR) can offer rich and reliable depth information. Compared to the pure-vision methods, the perception performance can be significantly improved by introducing new sensors. As a simple and efficient method for multi-modal fusion detection, BEVFusion~\cite{liu2023bevfusion} processes the inputs of LiDAR and camera separately, while mapping their features to a unified BEV space for fusion. Camera Radar Net (CRN)~\cite{kim2023crn} also applies an analogous fusion mechanism with different fuser modules, which performs well on the nuScenes dataset~\cite{nuscenes}. 
    

    
Despite those merits, some of these paradigms~\cite{liu2023bevfusion, kim2023craft} also face important limitations. Firstly, the fusers depend mostly on one modality rather than fully fusing two modalities. Secondly, LiDAR is sensitive to weather conditions~\cite{2021lidar} and is often costly, which limits its application in autonomous driving.


This paper proposes a 3D detection method based on 4D radar-camera fusion. Our method, formally termed ZFusion, adopts the advanced 4D radar and camera as two different sensors. 4D radar can provide 3D point cloud information (e.g., $x$-$y$-$z$ coordinate and radial velocity $v_r$). Due to its relatively low-cost advantages and strong adaption to all weather conditions~\cite{NEURIPS2022_185fdf62}, 4D radar has great potential to serve as an alternative to LiDAR. However, 4D radar has a much sparser point cloud than LiDAR, which contains much less information. In fact, the sparse point cloud of 4D radar has posed great challenges to effectively fusing it with the vision modality.
    
The pipeline of ZFusion is specified in Figure~\ref{fig:pipeline}, with a specific overall design of the fuser shown in Figure~\ref{fig:ZFusion}. The fuser aims to solve the difficulties during the fusion process. First, to catch and extract features optimally, we use the DCA mechanism to make local interaction of two modalities. Secondly, the information from 4D radar is sparse but highly credible, while the information from cameras is rich but relatively redundant, they are seriously unbalanced, our method can effectively balance the information of two modalities by sequentially combining the above modules but through different input orders. Finally, we consider multiple levels in the macro architecture to ensure more efficient fusion at different scales.


The main contributions of this work are as follows:

\begin{itemize}
    \item We propose a novel method called ZFusion for 4D radar-camera fusion, which has attractive performance in object perception. Specifically, experiments on the VoD dataset show that ZFusion achieved the \textit{state-of-the-art} $74.38\%$ mAP (mean average precision) in RoI (region of interest), while showing competitive mAP in EA (entire area) compared to baselines. Due to the stable performance of 4D radar in all-weather conditions, this method is subject to be extended to more complex conditions.

    \item We design an effective fusion block called DDCA. Owing to the merit of Transformer architecture, DDCA can effectively fuse information in our case, with great advantages over the convolutional-based fuser. Interestingly, by properly arranging the model, our Transformer block could well eliminate the negative impact brought by different query orders of modalities in single-direction attention-based fusion. For its strong ability to balance the multi-modal information, this block has the potential to be extended to other multi-modal fusion tasks, as a general thought.

    \item For image branch, we test and analyze the effect of depth supervision using 4D radar, and propose a new view transformation module suitable for both fusion and pure-vision task.
\end{itemize}

The rest of this paper is organized as follows. In Section~\ref{sec:related_work}, we introduce related works that inspired us. In Section~\ref{sec:methodology}, we provide the overall framework of ZFusion and the Transformer-based DDCA fusion block in detail. In Section~\ref{sec:result}, we present sufficient experimental results to show the performance of the proposed method, including the setup of experiments, ablation studies of different modules and relevant analyses. Finally, conclusion remarks and future works are given in Section~\ref{sec:conclusion}.

\section{Related Work}
\label{sec:related_work}

\subsection{LiDAR-Based 3D Object Detection}
LiDAR provides accurate depth information for autonomous driving. Most of the existing 3D detection approaches via LiDAR are point- or voxel-based. For example, point-based 3D detectors~\citep{yang20203dssd, shi2019pointrcnn} usually process the point clouds via PointNet~\citep{qi2017pointnet}, PointNet++~\citep{qi2017pointnet++} or PillarNet~\citep{shi2022pillarnet}, while solving the problem of disorder and non-uniformity of point clouds. VoxelNet~\citep{zhou2018voxelnet} and SECOND~\citep{yan2018second} divide the 3D space into regular voxels for 3D object detection. PointPillars~\citep{lang2019pointpillars} uses pillars, instead of voxels, to save computational resources and convert point clouds to pseudo-images.~\citet{yang2019std} and~\citet{shi2020pv} combine the advantages of both point- and voxel-based methods to improve the accuracy of 3D object detection. ~\citet{yin2021center} propose a LiDAR-based 3D object detection method based on a two-stage detection-optimization approach. Comparable results are shown in \ref{tab:modalities}.
    
\subsection{Camera-Based 3D Object Detection}
Compared with LiDAR solutions, camera-only solutions clearly have a cost advantage. Based on FCOS~\citep{9010746}, FCOS3D~\citep{wang2021fcos3d} decouples 2D and 3D features in the detection head for 3D perception. A series of work~\citep{wang2022detr3d, liu2022petr, chen2022graph} are built upon DETR~\citep{carion2020end}. Specifically, DETR3D extends DETR to a 3D detector by designing 3D-to-2D learnable object queries. 
    
Another stream of research~\citep{philion2020lift, roddick2018orthographic, xie2022m, reading2021categorical} obtains BEV features via depth prediction networks and feature converters. BEVDepth~\citep{bevdepth} also enriches the current frame information with point-based temporal fusion. BEVDet4D~\citep{huang2022bevdet4d} extends BEVDet~\citep{huang2021bevdet} by preserving the intermediate BEV features of past frames, where the increase of computational load is negligible. BEVFormer and its upgraded version use the spatio-temporal Transformer to fuse features of six cameras from different timestamps.
    
\subsection{Multi-Sensor Fusion}
Multi-Sensor Fusion is widely regarded as a solution to reach a performance-cost tradeoff. TransFusion~\citep{bai2022transfusion} improves the LiDAR-camera association mechanism to handle inferior image conditions. Deep Continuous Fusion~\citep{liang2018deep} and DeepFusion~\citep{li2022deepfusion} decorate the information captured by LiDAR and camera at the feature level. BEVFusion unifies multi-modal features in the BEV representation space. 
    
Considering the cost and robustness of sensors, cameras, and radars are adopted by more and more autonomous driving systems. For example, RC-PDA~\citep{long2021radar} proposes pixel depth association to find the one-to-many mapping between radar point and image pixels. CRAFT~\citep{kim2023craft} establishes a soft association between image proposal and radar point clouds in the polar-coordinate system. CRN utilizes 3D radar to assist view transformation and uses DCA for multi-modal feature fusion. RCFusion~\citep{rcfusion} designs an interactive attention module that fully utilizes the BEV features of both modalities. RCBEVDet~\citep{Lin_2024_CVPR} proposes a dual-stream radar backbone with both point-and Transformer-based blocks.

\begin{figure*}[ht]
    \begin{center}
    \centerline{\includegraphics[width= \linewidth]{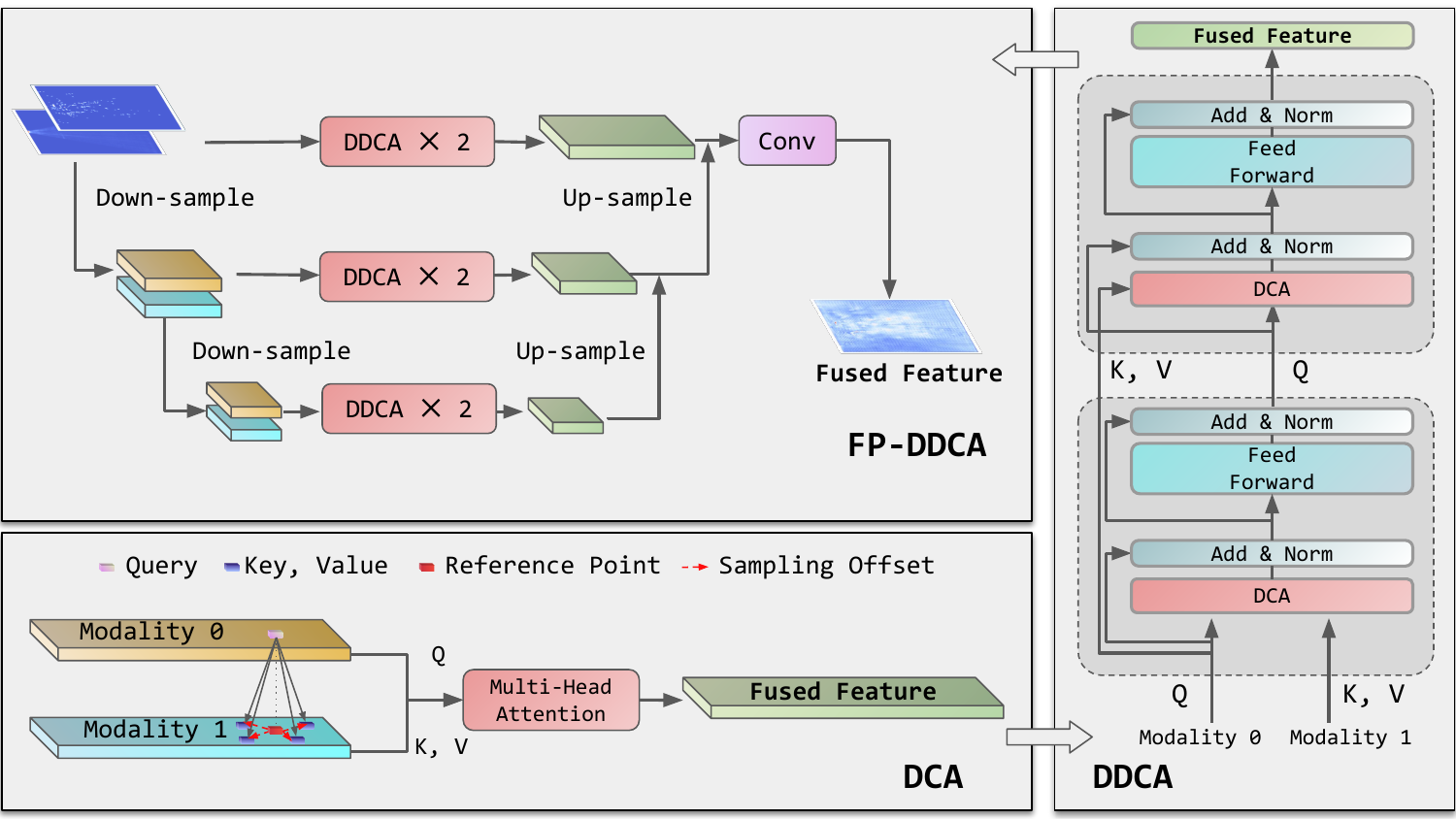}}
    \caption{Architecture of ZFusion, including the DCA mechanism (left-bottom), DDCA block (right), and FP-DDCA fuser (left-top).}
    \label{fig:ZFusion}
    \end{center} 
\end{figure*}

\section{Methodology}
\label{sec:methodology}

In this section, we introduce the details of ZFusion, 
including the fuser and the view transformation module. The main pipeline is given in Figure~\ref{fig:pipeline}, and the overall architecture of our proposed fuser is given in Figure~\ref{fig:ZFusion}.

\subsection{Fuser}

\subsubsection{The DCA Mechanism}

The convolution module is applied as the core of the fusion step in some works. Although convolution has many advantages, the fusion result is more like a simple superposition of two modalities, as illustrated in Figure~\ref{fig:feature_maps}.

As another mainstream fusion strategy, the multi-head attention mechanism is often believed to have the ability to better exploit the relationship between different features~\cite{NIPS2017_attention}. Based on this mechanism, Transformer has shown promising performance in multi-modal fusion~\cite{multimodaltransformer, mmtranssurvey}. The traditional Transformer requires a context to be as long as possible in natural language processing (NLP) tasks, termed ``global attention''. In computer vision and multi-modal fusion fields, considering the huge computing consumption of global attention mechanism, local attention is often believed to be a more efficient choice; see, e.g., Deformable Attention~\cite{Xia_2022_CVPR} and Deformable-DETR~\cite{detr2021}.

In real-world scenarios, establishing global relationship between the two modalities may not be necessary. To avoid both the non-ideal performance of convolution modules and $\mathcal{O}(n^2)$ computing consumption of global attention mechanisms, in this paper, we design a multi-modal fuser based on local attention mechanism. With the help of learnable offsets, this mechanism is able to effectively extract the relationship between different modalities in the unified BEV space. The key mechanism is called DCA, which consists of the following steps:
    
\begin{itemize}
    \item First, after aligning the features of two modalities, set the features of one modality as the query and map its central coordinate to the other modality as the reference point. 
    \item Next, the sampling offsets (i.e., relative positions between the reference point and sampling points) are learned for one modality.
    \item Then, set the feature from one modality as query (Q). According to the learnable sampling offsets, we get the features from other modalities (through bilinear interpolation) as keys (K's) and values (V's).
    \item Finally, apply the multi-head attention (MHA) to Q, K, and V to obtain the output feature. Formally, 
    \begin{align}
        & \hspace{-8.5 mm} \textrm{DCA}(Q_{i}, K_{1 - i}, V_{1 - i}) \nonumber \\
    = &~~\textrm{Concat}(head_1,\cdots, head_H)W^O,~~i \in \{ 0, 1\}
    \end{align}
    
    where
    \begin{align}
    head_h 
    = &~\textrm{DeformableAttn}_h (z_{q, i}, p_{q, i}, x_{1 - i})  \\
    = & \sum\limits_{n=1}^{N} A_{h, q, n, i \rightarrow 1 - i} \cdot [W_{h, i} \ x_{1 - i}(p'_{h, q, n, i \rightarrow 1 - i})],\nonumber \\
    & \hspace{-14mm} p'_{h, q, n, i \rightarrow 1 - i} = p_{q, i} + \Delta p_{h, q, n, i \rightarrow 1 - i},~~h = 1, \cdots, H,\nonumber 
    \end{align}


    $i$ refers to the index of modality, which takes values from $\{0,1\}$, $q$ denotes the index of a reference point, $H$ indicates the number of heads, and $N$ is the number of sampling points. $z_{q, i}$ is the $q$-th query feature of the $i$-th modality, $p_{q, i}$ is the coordinate position of the query feature, $x_{1 - i}$ is total feature of other modality in BEV space, $\Delta p$ indicates the learnable sampling offsets (relative position) between the reference point and sampling points (from one modality to the other).
\end{itemize}


\subsubsection{The DDCA Transformer Block}

From the perspective of input data, the intuition of this task is to balance the information from (sparse) radar point cloud and (dense) image. According to the toy fusion test, the result of a single DCA is similar to the input features of one modality (related to the input order) and shows an unsatisfactory result.

Our proposed solution is surprisingly simple. Based on the DCA mechanism described above, we repeatedly fuse different modalities but in different input orders. The benefit of this solution is not only a better balance of information between two modalities but also a powerful elimination of the undesirable effects of input order. The relevant experimental results are shown in Table~\ref{tab:modalities}. In this Transformer block, we use different modal features as Q, K, and V, respectively, and fuse them interchangeably in two steps. Specifically,
\begin{itemize}
    \item First, we set one modality as the Q while using the other as the K and V. 
    \item Second, we set the fusion modality as the Q and the first modality as the K and V. 
\end{itemize}


This fusion approach is easy to implement and works well. To the best of our knowledge, this is the first time this idea has been applied to feature fusion in BEV space, and it has the potential to be extended to other multi-modal fusion tasks. In addition, this block can effectively eliminate the influence brought by the modality interaction order. We therefore refer to this fusion block as the DDCA block. Except for the attention mechanism itself and different inputs, the overall architecture of the DDCA block mostly follows the architecture of the classic Transformer.

\subsubsection{The FP-DDCA Fuser}

In downstream tasks, different types of objects generally have different sizes, which is obvious in traffic scenarios. Naturally, we want to enhance modeling on multi-scale consideration, thereby improving the model's ability to extract multi-scale features to achieve better overall performances. To this end, the overall macro architecture of the fuser, termed Feature Pyramid-DDCA (FP-DDCA), is designed.

The feature pyramid architecture, represented by the well-known UNet~\cite{unet}, has significantly impacted the field of computer vision~\cite{fpn2017cvpr}. Its intuition lies in enhancing the learning performance by extracting features in different scales. In BEV space, our proposed architecture has similar considerations, in which features of different scales are obtained through downsampling before fusion. 

Specifically, for the input multi-modal features, the features are first downsampled by 1, 2 and 3 times, each followed by a DDCA block, and then upsampled by 1, 2 and 3 times, respectively. Finally, we output the fused feature through a convolutional layer. The architecture of the FP-DDCA fuser is specified in the left-top block of Figure~\ref{fig:ZFusion}. Using the pyramid architecture can bring some benefits, which will be seen in the experiment results section

Unlike the multi-scale deformable attention (MSDA) mechanism, which is widely used in the field of computer vision, we apply only single-scale deformable attention (e.g., $scale = 1$) module in each of our DCA mechanism, and use down/upsample before and after DDCA blocks to realize the multi-scale mechanism. The key difference between MSDA and our FP-DDCA is that in FP-DDCA, the DDCA blocks can balance the input features in each scale. 
To the best of our knowledge, this idea has been applied to the BEV space for the first time, which is easy to implement and achieves good results. The corresponding results are shown in Table~\ref{tab:FP}.

\subsection{View Transformation}

As the neck (e.g., view transformation module) of the image branch, LSS generates the camera's BEV features by mapping image features explicitly to BEV space. Vanilla LSS only uses the image feature extracted by the backbone to generate features distributed on different depths.

\begin{equation}
\begin{aligned}
    depth & = {\rm DepthNet}(x),   \\
    depth &\leftarrow {\rm Softmax}(depth),\\
    x_{\rm LSS} &= x \otimes depth,
\end{aligned}
\end{equation}

where $x$ is the image feature extracted by the backbone.

Generally, it is easy to realize that if the point cloud information mapped on the image is effectively used, the performance of the depth estimation module can be improved. Some works, such as BEVDepth, make good use of LiDAR point cloud space information in the image branch during both the training and inference phases. It is also possible to integrate space information from point clouds to LSS, which we refer to as depth supervision. For example, a Depth-supervised variant of LSS could get $x_{\rm LSS}$ by

\begin{equation}
\begin{aligned}
    x &\leftarrow {\rm Concat}([d, x]), \\
    x &\leftarrow {\rm DepthNet}(x),   \\
    depth &= {\rm Softmax}(x[:D]), \\
    x_{\rm LSS} &= x[D:] \otimes depth,
\end{aligned}
\end{equation}
$d$ is the feature extracted by a transformation network and represents the point cloud information mapped on the image.

\begin{figure}[ht]
    \begin{center}
    \centerline{\includegraphics[width=\linewidth]{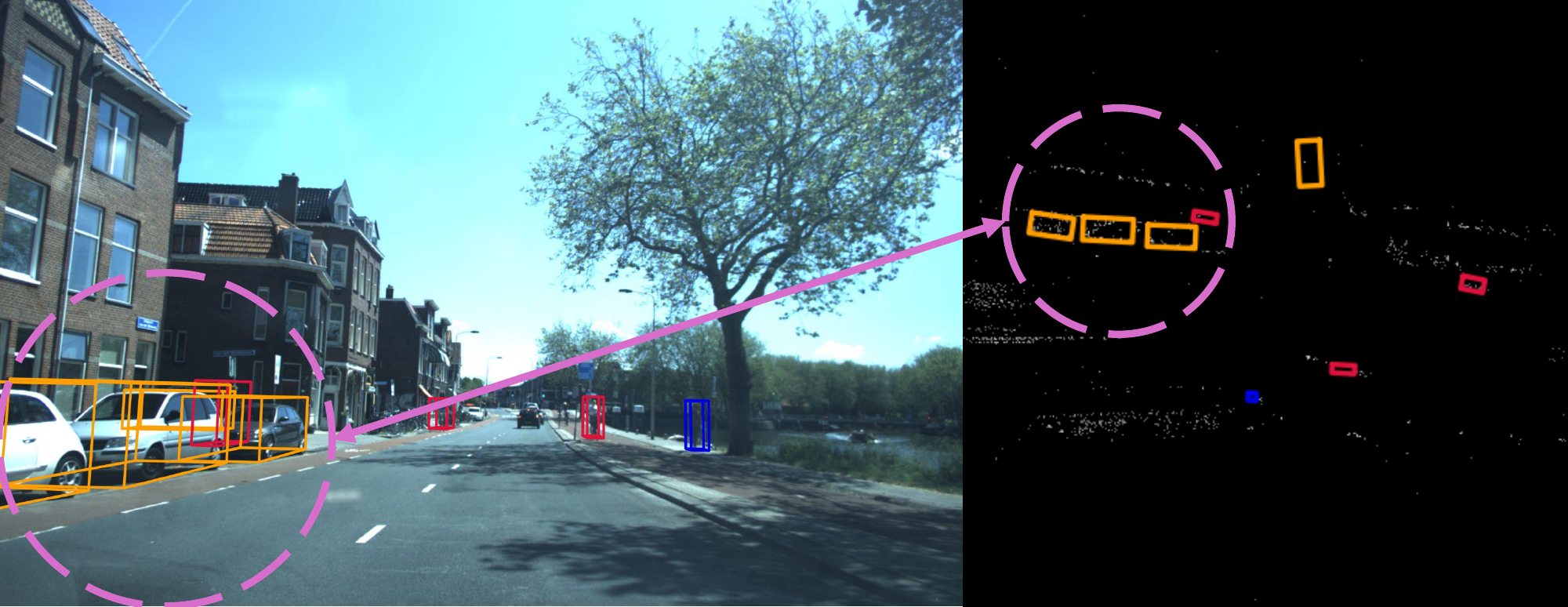}}
    \caption{4D radar can obtain some 'X-Ray vision' point clouds, shown in BEV space.}
    \label{fig:xray}
    \end{center}
\end{figure}

However, given the current input, we cannot directly use the Depth-supervised LSS variant here. Essentially, it is related to the physical properties of 4D radar. The relatively large wavelength gives 4D radar ``X-ray vision'' to detect objects behind obstruction; a visualization of this phenomenon is given in Figure~\ref{fig:xray} \footnote{Lots of reasonable obscured objects have already appeared in our testing results.}. On the other hand, it also makes point cloud-supervised depth estimation less reliable than LiDAR, since some of the point clouds from the 4D radar contain ``X-ray vision'' and some noise, ``sees'' something different from camera, leading to incorrect supervision. Therefore, it makes logical sense to retain the 4D radar's position information in the overall space but not rely on the supervision from the point cloud for depth estimation.

\begin{equation}
\begin{aligned}
    depth & = {\rm DepthNet}(x),   \\
    depth &\leftarrow {\rm Softmax}(depth),\\
    x &\leftarrow {\rm ContextNet}(x),\\
    x_{\rm LSS} &= x \otimes depth,
\end{aligned}
\end{equation}

Based on the above considerations, we utilize a new variant of LSS as an alternative. This LSS variant does not use any form of point cloud information, clearly separates estimated context and depth features, and achieves better performance in our scenario than other strategies, corresponding experimental results are shown in Table~\ref{tab:depth}.

\section{Experimental Results}
\label{sec:result}

    \begin{figure*}[ht]
    \begin{center}
    \centerline{\includegraphics[width=0.5\linewidth]{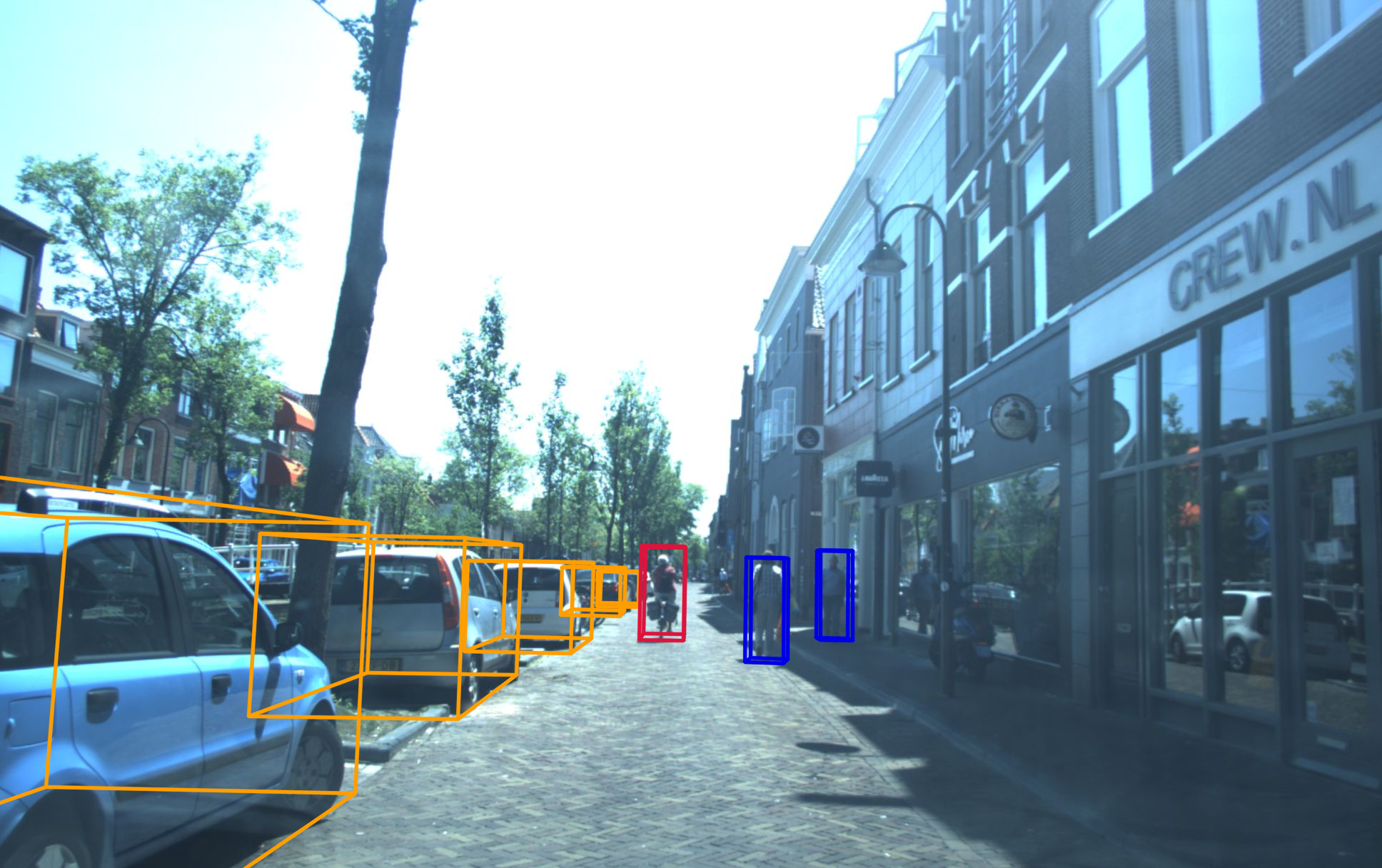}\includegraphics[width=0.5\linewidth]{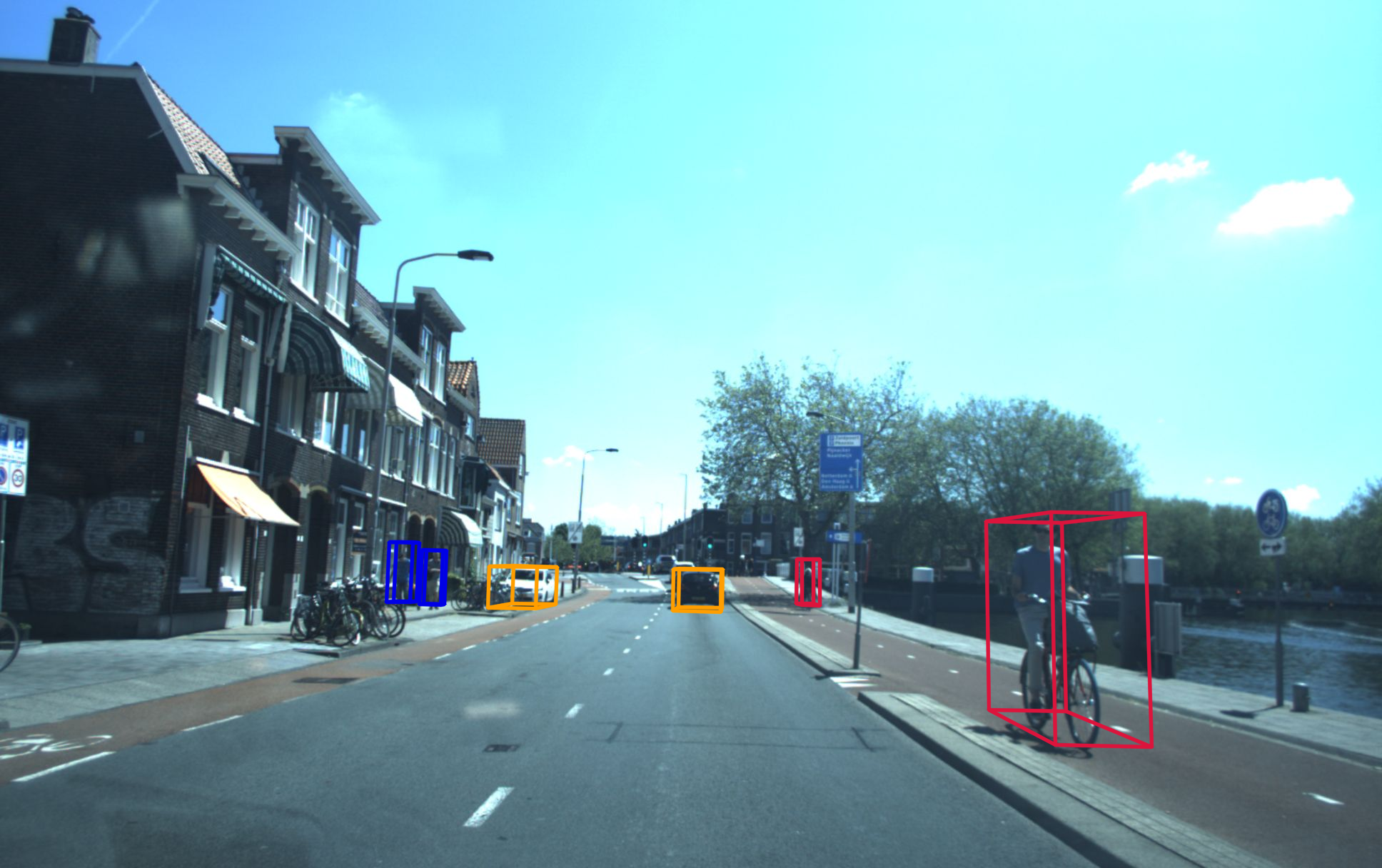}}
    \caption{Typical visualizations in VoD test set, ID: 0262 (left) and 4942 (right)}
    \label{VoD_test}
    \end{center}
    \end{figure*}

\begin{table*}[ht]
    \centering
    \resizebox{\textwidth}{!}{%
    \begin{tabular}{ccccccccccc}
    \toprule
    \multirow{2}{*}{Method} & \multirow{2}{*}{Modality} & \multicolumn{4}{c}{AP in the Entire Area (\%)} & \multicolumn{4}{c}{AP in the Region of Interest (\%)} & \multirow{2}{*}{FPS} \\
                            &                           & Car      & Pedestrian    & Cyclist   & mAP     & Car        & Pedestrian     & Cyclist     & mAP       &                      \\ \midrule
    PointPillars (CVPR 2019)           & R                         & 37.06    & 35.04         & 63.44     & 45.18   & 70.15      & 47.22          & 85.07       & 67.48     & N/A                  \\
    SMURF (TIV 2023)                  & R                         & 42.31    & 39.09         & \underline{71.50}& 50.97   & 71.74      & 50.54          & 86.87       & 69.72     & 23.1                 \\
    RCFusion (TIM 2023)               & C+R                       & 41.70    & 38.95         & 68.31     & 49.65   & 71.87      & 47.50          & 88.33       & 69.23     & 4.7--10.8            \\
    BEVFusion (ICRA 2023)              & C+R                       & 37.85    & \underline{40.96}& 68.95     & 49.25   & 70.21      & 45.86          & \underline{89.48}& 68.52     & 7.1                  \\
    BEVFusion\tablefootnote{Our implementation of BEVFusion.} (ICRA 2023)            & C+R                       & 37.74    & 34.05         & 64.63     & 45.47   & 70.99      & 43.63          & 86.41       & 67.01     &14.8                   \\
    RCBEVDet (CVPR 2024)               & C+R                       & 40.63    & 38.86         & 70.48     & 49.99   & 72.48      & 49.89          & 87.01       & 69.80     & 21--28                  \\
    LXL (TIV 2024)                    & C+R                       & 42.33    & \textbf{49.48}         & \textbf{77.12}     & \textbf{56.31}   & 72.18      & \textbf{58.30}          & 88.31       & 72.93     & 6.1                  \\ \midrule
    ZFusion                     & C+R                       & \textbf{44.77}   & 40.02         & 68.61     & 51.14   & \textbf{80.44}      & 52.68          & \textbf{90.01}       & \textbf{74.38}     &8.8             \\ 
    ZFusion (ResNeXt-50)         & C+R                       &\underline{43.89}      &39.48           &70.46       &\underline{51.28}&\underline{79.51}&\underline{52.95}&86.37         &\underline{72.94}&11.2                    \\
    \bottomrule         
    \end{tabular}%
    }
    \caption{Per-class APs of different models in the entire area and Region of Interest. ``R'', ``C'' and ``L'' denote radar, camera and LiDAR modality respectively. All the baseline APs are inherited from LXL~\cite{lxl}. The default version of ZFusion uses ResNet-101 as the image backbone.}
    \label{tab:results}
\end{table*}

\subsection{Dataset}
As a well-known public dataset, the VoD dataset~\cite{apalffy2022vod} provides complex, urban traffic scenes with 13 object classes including $car$, $pedestrain$, and $cyclist$. An accurate 3D bounding box annotation is provided for every object, including $\{ x, y, z, w, l, h, \theta\}$, where $\theta$ refers to the rotation of objects around the LiDAR’s negative vertical ($-z$) axis~\cite{apalffy2022vod}. For any timestamp, VoD provides the stacked point clouds of the previous 1, 3, and 5 frames, which are aligned according to the displacement of car. Compared with the well-known dataset nuScenes, the VoD dataset additionally provides 4D point cloud (i.e., $\{x, y, z, v_r\}$ information) by a 4D radar with a frequency of approximately $13$Hz, where $v_r$ refers to the absolute (i.e. ego motion compensated) radial velocity of the point, while nuScenes provides no $z$-axis information of radar point clouds.

According to the VoD official split, training data of $5,139$ frames and validation data of $1,296$ frames are used to train and test the model, respectively. In the experiments, we consider $car$, $pedestrian$, and $cyclist$, the three main object categories measured by the official VoD metrics. Following~\cite{liu2023bevfusion}, the images are augmented by random rotations, flips, and cuts. The radar points are augmented by random rotation, flips, and re-scaling. All augmentation matrices are recorded to keep consistency between data (or features) of two modalities. 

\subsection{Implementation Details}


\noindent \textbf{Details of the model.} The model is standardly composed of the backbone, neck, and head components. ResNet/ResNeXt~\cite{He2015DeepRL, Xie2016} and a variant of LSS~\cite{philion2020lift} are used as the backbone and neck for the image branch, respectively.
For the radar branch, the point cloud range (PCR) in the radar coordinate is set to 
    \begin{align*}
        D_{\text{PCR}} = \{(x, y, z) \ | & \ 0 < x < 51.2\text{m}, \\  -25.6\text{m} < y < 25.6\text{m}, 
        & \ -3\text{m} <z<2\text{m}\}
    \end{align*}
as suggested by the official guidelines of the VoD dataset. We use VoxelNet~\cite{zhou2018voxelnet} as the backbone, where the size of voxel is set to be $0.16\text{m}\times 0.16\text{m} \times 0.24\text{m}$.
For the fuser, the number $H$ of heads is set to $8$ and the number $N$ of sampling points is set to $32$. 

\noindent \textbf{Training and test setups.} 
We train and test our model on a Linux server with 8 $\times$ NVIDIA RTX A6000 GPUs with $48$GB memory each. The code is built on the MMCV library. We train our model for about 100 epochs with the AdamW optimizer~\cite{Loshchilov2017DecoupledWD}. By simply searching from range $[10^{-5}, 10^{-3}]$, we choose the learning rate as $10^{-4}$. For each setting, our experiments are conducted only once with randomness, and the results are recorded. 

\noindent \textbf{Metrics.}
We use the official VoD evaluation scripts, which include per-class average precision (AP) for the entire annotated area and per-class AP for the driving corridor, i.e., AP in Region of Interest (RoI AP). The range of the driving corridor is defined in the camera coordinate as $$D_{RoI} = \{(x, y, z) \ | \ 0 < x < 25\text{m}, \ -4\text{m} < y < 4\text{m}\}.$$ The AP is calculated based on Intersection over Union (IoU). A predicted box is defined as a true positive prediction if the IoU between itself and a ground truth box is above the IoU threshold of $0.5$, $0.25$, and $0.25$ for $car$, $pedestrian$, and $cyclist$, respectively. The thresholds are strictly following the official evaluation requirements.

\subsection{Results on VoD Dataset}

\begin{figure*}[ht]
    \centering
    \subcaptionbox{Camera}{\includegraphics[width=0.245\linewidth]{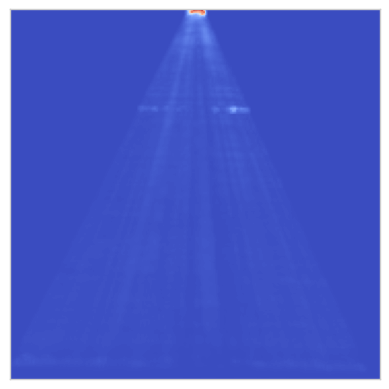}}
    \subcaptionbox{4D radar}{\includegraphics[width=0.245\linewidth]{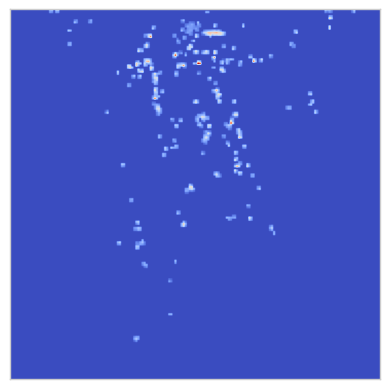}}
    \subcaptionbox{Fused by Convolution}{\includegraphics[width=0.245\linewidth]{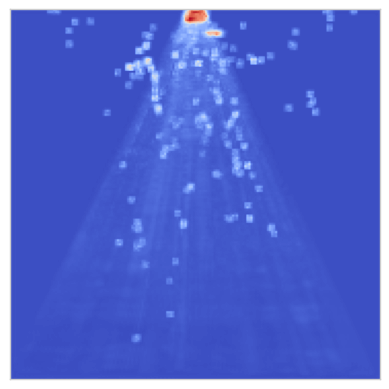}}
    \subcaptionbox{Fused by FP-DDCA}{\includegraphics[width=0.245\linewidth]{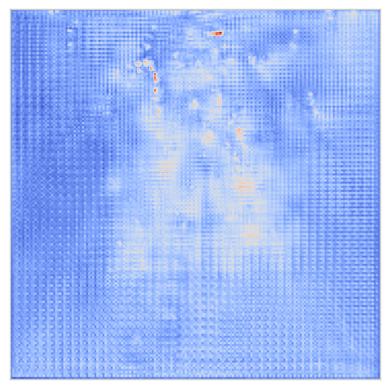}}
    
    \caption{Feature maps of different modalities and feature maps fused by Convolution and FP-DDCA.}
    \label{fig:feature_maps}
\end{figure*}

\begin{table*}[tb]
\centering
\begin{tabular}{ccccccccccc}
\toprule
\multirow{2}{*}{Modality} & \multirow{2}{*}{Method} & \multicolumn{4}{c}{AP in the Entire Area (\%)} & \multicolumn{4}{c}{AP in the Region of Interest (\%)} \\
                        &                           & Car      & Pedestrian    & Cyclist   & mAP     & Car        & Pedestrian     & Cyclist     & mAP       &                      \\ \midrule
R & SMURF                 & 42.31    & 39.09         & \textbf{71.50}     & 50.97   & 71.74      & 50.54          & 86.87       & 69.72      \\
C & CenterNet             &  25.94   & 29.76     &   34.98   &  30.23  &   50.87    &     36.81      &    57.42    &   48.37   \\
C+L & MVX-Net             &  \textbf{78.25}   &   \textbf{51.76}       &   52.26   &  60.76  &   \textbf{90.62}    &    \textbf{68.16}       &    72.03    &   \textbf{76.94}   \\
L & IA-SSD             &  77.29   &   32.18       &   57.11   &  55.53  &   N/A    &    N/A      &    N/A    &   N/A   \\
\midrule
C+R &   ZFusion                    &  43.89    & 39.48         & 70.46     & 51.28   & 79.51      & 52.95          & 86.37       & 72.94   \\
R+C &   ZFusion                 &  41.60    &     42.29     &    71.23  &  51.71  &   71.94    &     55.12      &    \textbf{89.70}    &  72.25    \\
\bottomrule         
\end{tabular}%

\caption{Per-class APs of different modalities, using ResNeXt-50 as image backbone. ``R+C'' means that the radar and camera features are set as modality $0$ and modality $1$ in the first DCA, respectively (see the right part in Figure~\ref{fig:pipeline}). APs of CenterNet~\cite{Zhou2019ObjectsAP} and ~\cite{iassd}
are inherited from~\citet{zhang2024tl} and~\citet{deng2023seeing}, respectively.}
\label{tab:modalities}
\end{table*}

\begin{table}[tb]
\centering
\begin{tabular}{ccc}
\toprule
Fusion Block in FP & mAP & RoI mAP\\ \midrule
DDCA         &        \textbf{51.14}                &          \textbf{74.38}                     \\
Convolution  &   47.49   &    68.84                  \\ \bottomrule
\end{tabular}
\caption{mAPs of different basic fusion blocks.}
\label{tab:DDCA}
\end{table}

In Table~\ref{tab:results}, we compare the APs of our proposed ZFusion with four well-known 4D radar-camera fusion methods, i.e., BEVFusion~\cite{liu2023bevfusion}, RCFusion~\cite{rcfusion}, RCBEVDet\cite{Lin_2024_CVPR}, LXL~\cite{lxl}, and two radar-only methods: SMURF~\cite{smurf} and PointPillars~\cite{lang2019pointpillars}. 

Among all the methods under test, ZFusion gains \textit{state-of-the-art} RoI mAP, with a competitive performance in the Entire Area. In terms of APs per class, ZFusion achieves the best \textit{car} AP, car RoI AP, and \textit{cyclist} RoI AP, increasing by $2.44\%$, $7.96\%$ and $0.53\%$ upon LXL, RCBEVDet and BEVFusion, respectively. We point out that compared to our implementation of BEVFusion in the VoD dataset, ZFusion increases by $5.67\%$ on mAP and $7.37\%$ on RoI mAP, respectively. This increase, partially owing to the fusion capability of FP-DDCA, will be discussed in detail in the ablation study.


We also report the inference speeds of all the methods in Table~\ref{tab:results}. Specifically, ZFusion and ZFusion (ResNeXt-50) get $8.8$ FPS and $11.2$ FPS, respectively, under PyTorch implementation. The inference speeds are reasonable and remain competitive among all R+C methods.

\subsection{Ablation Study}

\subsubsection{Effects of Different Modalities}

The results of different sensor combinations (radar-only, camera-only, LiDAR+camera, radar+camera) and corresponding methods are listed in Table \ref{tab:modalities}. The intuition is that the modalities with LiDAR gain higher APs than those with 4D radar on most objects. Still, it is also surprising that the radar+camera modality accounts for $85.11\%$ of the LiDAR+camera mAP, and even exceeds the methods involved with LiDAR in \textit{cyclists}. 
Meanwhile, compared with the camera-only modality, the radar+camera modality gains $21.48\%$ and $23.88\%$ increases on mAP and RoI mAP, respectively. Through our method, radar+camera combination is able to approach the effect of using the LiDAR modality, while significantly exceeding the overall performance of the camera-only modality.



\begin{table*}[tb]
\centering
\begin{tabular}{ccccccccc}
\toprule
\multirow{2}{*}{FP Layers} & \multicolumn{4}{c}{AP in the Entire Area (\%)} & \multicolumn{4}{c}{AP in the Region of Interest (\%)} \\
                          & Car      & Pedestrian    & Cyclist   & mAP     & Car        & Pedestrian     & Cyclist     & mAP       \\ \midrule
1 (no FP)                         &  43.68   &    38.59      &   66.26   &  49.51  &   71.53    &     50.66      &    84.79    &   68.99   \\
2                       &  43.48   &     \textbf{39.98}     &   66.99   &  50.15  &    71.14   &     51.55      &   \textbf{86.43}     &   69.71   \\
3                     &\textbf{43.89}      &39.48           &\textbf{70.46}       &\textbf{51.28}     &\textbf{79.51}        &\textbf{52.95}            &86.37         &\textbf{72.94}   \\ \bottomrule
\end{tabular}
\caption{Per-class APs of different layers of FP, using ZFusion (ResNeXt-50).}
\label{tab:FP}
\end{table*}

\subsubsection{Effects of DDCA Module}

We conduct experiments using the DDCA blocks and convolution in FP, respectively, to test the effects of the DDCA block. The results in Table~\ref{tab:DDCA} show that DDCA works obviously better, gaining $3.65\%$ and $5.54\%$ increases in both mAP and RoI mAP. We also plot the feature maps of both methods. It can be observed in Figure~\ref{fig:feature_maps} that the features fused by DDCA show more balance, while the features fused by convolution make poor use of camera modality.



The effects of the interaction order in DDCA are tested and the APs are reported in Table~\ref{tab:modalities}. Here, ``R+C'' means that the radar and camera features are set as modality 0 and modality 1 in the first DCA, respectively (see the right part in Figure~\ref{fig:ZFusion}), and vice versa. The results show no significant difference between different interaction orders, demonstrating that DDCA is a balanced fusion module between the two modalities.





\subsubsection{Effects of FP}

To capture multi-scale features, many works, including UNet~\cite{unet}, FPN~\cite{fpn2017cvpr}, and Deformable DETR~\cite{detr2021}, utilize the idea of a pyramid structure. Following that, we also propose a feature pyramid (FP) architecture in ZFusion. To analyze the contribution of FP, we conduct experiments using 1 layer (i.e., no FP), 2 layers, and 3 layers of FP, where each layer contains 2 DDCA blocks.
The results are shown in Table \ref{tab:FP}. The \textit{car} RoI AP, \textit{cyclist} RoI AP, and RoI mAP improve by $7.98\%$, $2.29\%$, and $3.95\%$, respectively, from no FP to 3-layer FP, which validates the advantage of FP using multi-level features. 

\subsubsection{Effects of View Transformation Module} 

To explore the effect of view transformation module under our sensor combination, we trained ZFusion using all three strategies, to test variants shown in Methodology. The results are listed in Table~\ref{tab:depth}, which shows that in our ``R+C'' sensor combination, depth supervision by 4D radar has negative effect, while the new variant we proposed achieved better results than the vanilla version.

\begin{table}[tb]
\centering
\begin{tabular}{cccc}
\toprule
Version & Supervision & mAP & RoI mAP \\ \midrule  
Depth-Supervised \tablefootnote{Supervised by 4D radar point cloud here.} &     \checkmark     &     48.75                   &      69.29                        \\
Vanilla &     $\times$     &     49.01                   &      72.10                        \\
Depth-Context &    $\times$      &       \textbf{51.14}                &          \textbf{74.38}                           \\ \bottomrule
\end{tabular}
\caption{mAPs of different view transformation module.}
\label{tab:depth}
\end{table}

\subsubsection{Effects of Point Cloud Density} The VoD dataset provides the stack of point clounds in previous 1 (no stack), 3, and 5 frames, compensated by displacements of ego cars across different frames. 

To explore the effect of point cloud density, we test the APs of different frame accumulation numbers. The results are listed in Table~\ref{tab:frames}. The RoI mAPs increase stably from $1$ frame to $5$ frames, while the mAPs are relatively close. When the objects are close to the ego car, considering objects' movement and noise of 4D radar, the stack of closer point clouds is generally more accurate. When objects are far away from the ego car or move fast, the stack may not be accurate. In addition, the image branch can also provide more semantic information in closer regions (e.g. RoI).


\begin{table}[tb]
\centering
\begin{tabular}{ccc}
\toprule
\# of Frames & mAP & RoI mAP \\ \midrule  
1         &     49.27&      68.42\\
3         &       48.49&          69.29\\
5         &       \textbf{51.28}&          \textbf{72.94}\\
\bottomrule
\end{tabular}
\caption{mAPs of different numbers of frames.}
\label{tab:frames}
\end{table}
\section{Conclusion}
\label{sec:conclusion}

As the key to autonomous driving, reliable 3D perception relies on strong cooperation between sensors~\cite{multisensorsurvey}. This paper proposes a novel method called ZFusion for 4D radar-camera fusion. With a new view transformation module, our method can effectively fuse the camera and 4D radar in the shared BEV space. Since our method performs fusion based on DCA mechanism between different modalities, 
which only pays attention to a small set of key sampling points around the corresponding coordinates, regardless of the spatial size of the feature maps. In addition, DDCA blocks are designed to balance information better. With an overall multi-scale architecture, a more sufficient fusion of multi-modal information can be ensured. Experimental results demonstrate that, compared to the baselines, ZFusion partially achieved \textit{state-of-the-art} performance in typical traffic scenarios, especially in the driving corridor (RoI). Considering that cameras and LiDARs do not work well in complex weather conditions, our proposed method is suitable for all-weather autonomous driving. Our results indicate that the performance of proposed 4D radar-camera fusion scheme greatly exceeds the camera-only scheme and is close to the high-cost LiDAR-based scheme, which can serve as an attractive solution for automatic driving perception. Also, we want to point out some directions worthy of future investigation, which include evaluating and improving overall performance in low-visibility weather and gaining robustness against partial hardware-missing cases.
    
    
    
\section*{Acknowledge}
\label{sec:acknowledge}



This research was jointly supported by the National Key R$\&$D Program of China (2024YFF0505601), the National Natural Science Foundation of China (62471147), the Open Projects Program of State Key Laboratory of Multimodal Artificial Intelligence Systems, and ZF (China) Investment Co., Ltd. The computations in this research were performed using the ZF computing servers and the CFFF platform of Fudan University. The authors sincerely thank the relevant departments for data collection, labeling, and computing resources.
{
    \small
    \bibliographystyle{ieeenat_fullname}
    \bibliography{main}
}


\end{document}